\def\paperTitle{Learn to Optimize Denoising Scores for 3D Generation: A Unified and Improved Diffusion Prior on NeRF and 3D Gaussian Splatting}

\def\authorBlock{
    Xiaofeng Yang\thanks{Equal contribution} $^1$ \qquad
    Yiwen Chen\footnotemark[1] $^1$ \qquad
    Cheng Chen$^1$ \qquad Chi Zhang$^1$ \\ \qquad
    Yi Xu$^2$  \qquad 
    Xulei Yang$^3$ \qquad
    Fayao Liu$^3$  \qquad
    Guosheng Lin\thanks{Corresponding Author} $^1$ \\ 
    $^1$ Nanyang Technological University \quad $^2$ OPPO US Research Center \quad \\
    $^3$ A*STAR, Singapore \quad \\
    {\tt\small \{yang.xiaofeng, gslin\}@ntu.edu.sg}
}

\newif\ifreview 
\newif\ifarxiv \newcommand{\arxiv}{\arxivtrue}
\newif\ifcamera 
\newif\ifrebuttal

\arxiv
\pdfoutput=1
\documentclass[10pt,twocolumn,letterpaper]{article}
\ifreview \usepackage[review]{cvpr} \fi
\ifarxiv \usepackage[pagenumbers]{cvpr} \fi
\ifrebuttal \usepackage[rebuttal]{cvpr} \fi
\ifcamera \usepackage{cvpr} \fi

\usepackage{graphicx}	
\usepackage{amsmath}	
\usepackage{amssymb}	
\usepackage{booktabs}
\usepackage{times}
\usepackage{microtype}
\usepackage{epsfig}
\usepackage[table,xcdraw,dvipsnames]{xcolor}
\usepackage{caption}
\usepackage{float}
\usepackage{placeins}
\usepackage{color, colortbl}
\usepackage{stfloats}
\usepackage{enumitem}
\usepackage{tabularx}
\usepackage{xstring}
\usepackage{multirow}
\usepackage{xspace}
\usepackage{url}
\usepackage{subcaption}
\usepackage{xcolor}
\usepackage[hang,flushmargin]{footmisc}

\ifcamera \usepackage[accsupp]{axessibility} \fi

% ...

\ifarxiv  \fi

\newcommand{\R}[1]{{%
    \textbf{%
        \ifstrequal{#1}{1}{\textcolor{red}{R#1}}{%
        \ifstrequal{#1}{2}{\textcolor{blue}{R#1}}{%
        \ifstrequal{#1}{3}{\textcolor{magenta}{R#1}}{%
        \ifstrequal{#1}{4}{\textcolor{teal}{R#1}}{%
                           \textcolor{cyan}{R#1}%
        }}}}%
    }%
}}

\newcommand{\zt}[0]{z_t}

\newcommand{\distributionp}[0]{ p_\phi(z;y)}

\newcommand{\scorep}[0]{ \triangledown_z\log p_\phi(z;y)}

\newcommand{\modelcondition}[0]{ \epsilon_\phi(\zt;y)}

\newcommand{\ldist}[0]{\mathcal{L}_{\text{SDS}}}
\newcommand{\ldistnorm}[0]{\mathcal{L}_{\text{SDS--norm}}}
\newcommand{\ldistref}[0]{\mathcal{L}_{\text{SDS--ref}}}
\newcommand{\ldistdds}[0]{\mathcal{L}_{\text{DDS}}}
\newcommand{\ldistvsd}[0]{\mathcal{L}_{\text{VSD}}}

\usepackage[linesnumbered,ruled,vlined]{algorithm2e}
\SetKwInput{KwInput}{Input}                
\SetKwInput{KwOutput}{Output}              

\usepackage{xr-hyper}

\makeatletter
\newcommand*{\addFileDependency}[1]{
  \typeout{(#1)}
  \@addtofilelist{#1}
  \IfFileExists{#1}{}{\typeout{No file #1.}}
}

\makeatother

\definecolor{cvprblue}{rgb}{0.21,0.49,0.74}
\usepackage[pagebackref,breaklinks,colorlinks,citecolor=cvprblue]{hyperref}
\usepackage[capitalize]{cleveref}
\crefname{section}{Sec.}{Secs.}
\crefname{table}{Table}{Tables}
\crefname{figure}{Fig.}{Figs.}

\begin{document}
\title{\paperTitle}
\author{\authorBlock}
\maketitle

\begin{abstract}
In this paper, we propose a unified framework aimed at enhancing the diffusion priors for 3D generation tasks. Despite the critical importance of these tasks, existing methodologies often struggle to generate high-caliber results. We begin by examining the inherent limitations in previous diffusion priors. We identify a divergence between the diffusion priors and the training procedures of diffusion models that substantially impairs the quality of 3D generation. To address this issue, we propose a novel, unified framework that iteratively optimizes both the 3D model and the diffusion prior. Leveraging the different learnable parameters of the diffusion prior, our approach offers multiple configurations, affording various trade-offs between performance and implementation complexity. Notably, our experimental results demonstrate that our method markedly surpasses existing techniques, establishing new state-of-the-art in the realm of text-to-3D generation. Furthermore, our approach exhibits impressive performance on both NeRF and the newly introduced 3D Gaussian Splatting backbones. Additionally, our framework yields insightful contributions to the understanding of recent score distillation methods, such as the VSD and DDS loss. Project Page and Code: \href{https://yangxiaofeng.github.io/demo_diffusion_prior}{Link}
\end{abstract}
\section{Introduction}
\label{sec:intro}
\begin{figure*}[ht]
\begin{center}
\includegraphics[trim=0 0 0 0,width=1\linewidth]{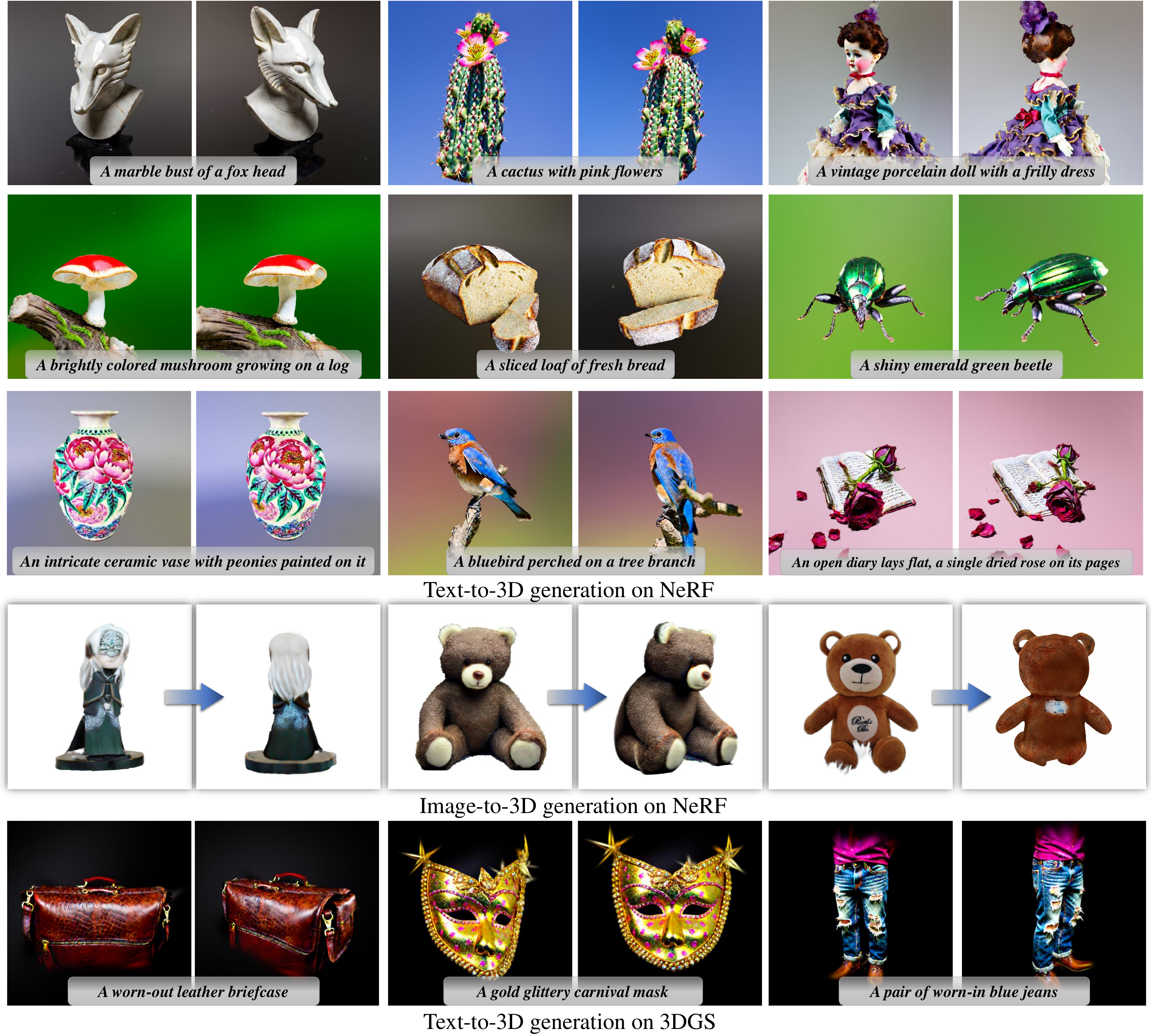}
	\caption{Examples of 3D generation results of LODS. Our proposed method is capable of generating 3D objects with exceptional fidelity, showcasing intricate details and remarkably accurate colors.}
	\label{fig:demo}
\end{center}
\vspace{-0.4cm}  
\end{figure*}
Diffusion models, a cutting-edge generative methodology, have garnered increasing research attention in recent years~\cite{ho2020denoising,song2020score,song2020denoising,dhariwal2021diffusion,nichol2022glide,ho2022cascaded,sahariaphotorealistic,ramesh2022hierarchical,rombach2022high}. Diffusion models trained on massive datasets comprising billions of images, demonstrate the capability to achieve high-fidelity text-guided image generation~\cite{nichol2022glide,sahariaphotorealistic,ramesh2022hierarchical,rombach2022high}. In addition to their proficiency in 2D image generation, various methods~\cite{hertz2023delta,poole2022dreamfusion,wang2023prolificdreamer} have been proposed to apply diffusion prior in a diverse range of optimization problems. These applications encompass conditional 2D generation~\cite{graikos2022diffusion}, 2D image editing~\cite{graikos2022diffusion,hertz2023delta}, 3D generation~\cite{poole2022dreamfusion,wang2023prolificdreamer} and 3D editing~\cite{haque2023instruct,yu2023edit}.  

Seminal work by Graikos \textit{et al.}~\cite{graikos2022diffusion} shows that the pretrained diffusion models combined with constraints can be used as a plug-and-play prior to solve multiple conditional generation problems. The milestone works, Dreamfusion~\cite{poole2022dreamfusion} and Score Jacobian Chaining~\cite{wang2023score}, demonstrate the importance of removing the U-Net Jacobian when calculating the diffusion prior and show great performance in text to 3D generation. Although the SDS loss proposed in Dreamfusion~\cite{poole2022dreamfusion} represents a breakthrough, it exhibits notable limitations. Firstly, the SDS loss often results in blurred generative images, and secondly, it struggles to capture the diversity of the data space. Recent research endeavors have sought to enhance the SDS loss from different perspectives. The DDS loss~\cite{hertz2023delta}, targeting the 2D image editing tasks, has demonstrated that the original SDS loss produces noisy score maps even on real images. To solve this issue, DDS loss proposes to remove the noise by taking the difference of the SDS losses between the current image and the source image. ProlificDreamer~\cite{wang2023prolificdreamer} proposes VSD loss to improve 3D generation quality by learning the variational distribution of the text prompt using LoRA~\cite{hu2021lora}. Despite these advancements, the underlying reasons for the sub-optimal performance of the SDS loss and strategies to rectify its weaknesses remain largely unexplored.

In this research, our primary objective is to enhance the quality of 3D generation by improving the diffusion priors. Our initial step involves a thorough investigation into the limitations of existing diffusion priors. We discover that the SDS loss employs classifier free guidance (CFG) in optimizing the 3D model. However, during the training phase of diffusion models, CFG is not utilized, leading to a discrepancy between the training and inference stages of diffusion models. This discrepancy results in the SDS loss directing the 3D model towards the CFG-oriented path rather than the original diffusion model path without CFG, culminating in suboptimal outcomes.

To address this issue, we introduce LODS (\textbf{L}earn to \textbf{O}ptimize \textbf{D}enoising \textbf{S}cores), a unified framework that iteratively optimizes both the 3D model and the SDS, aligning the 3D model more closely with the distribution score learned by the original diffusion model. LODS extends the SDS by incorporating additional learnable parameters. Within this framework, we examine two types of learnable parameters: the learnable null embedding and the low-rank model parameters. These methods offer flexibility, allowing for various trade-offs between generation performance and implementation complexity. Notably, our embedding-based LODS method can be implemented with the addition of approximately twenty lines of code. Despite its simplicity, it significantly enhances generation performance.

Our 3D experiments are carried out on text-guided 3D generation and image-guided 3D generation. We show that our proposed methods greatly improve the 3D generation quality and establish new state-of-the-art performances on text-to-3D benchmarks, surpassing the recently introduced VSD loss. While our primary emphasis revolves around addressing 3D challenges, we also provide qualitative evidence of our method's capacity to produce high-quality 2D generation and editing outcomes. Some 3D generation results of our proposed diffusion prior can be found in Fig.~\ref{fig:demo}.

To summarize our contributions:
\begin{itemize}    
    
    \item We study the problems of previously introduced diffusion priors and propose a unified framework to enhance the diffusion prior by learning to optimize the denoising scores.
    
    \item Our approach can be applied to a wide spectrum of use cases and consistently yields substantial performance improvements.
    
    \item We show in experiments that our method achieves new state-of-the-art performance in 3D generation benchmark T3Bench, significantly outperforming other contemporary methods. 
\end{itemize}
\section{Related Work}
\label{sec:related}
\subsection{Diffusion Models}
\label{sec:relateddiffusion}
There are several interpretations of diffusion models, either as score functions~\cite{song2019generative,song2020improved,song2020score} or from the perspective of the diffusion process~\cite{ho2020denoising,song2020denoising}. Based on the score function interpretations~\cite{song2019generative,song2020improved,song2020score}, the diffusion model learns the score function of the distribution $\distributionp$ by:
\begin{align}
 \mathbb{E}_{p_\phi(z)}[\parallel \modelcondition - \scorep \parallel_2^2],
\label{eq:energy_score}
\end{align}
where $\modelcondition$ is the diffusion model. During training, the score function $\scorep$ is defined by the noise adding process. After the diffusion model is learned, it is possible to generate data from a random initial point of the data space by following the direction of the diffusion model prediction. We can roughly say $\modelcondition \approx \scorep$ on a learned diffusion model.

\textbf{Classifier Free Guidance}~\cite{ho2021classifier} (CFG) is a method to generate conditional content without an external classifier. Specifically, the classifier free guidance is defined as: 
\begin{align}
\hat\epsilon_\phi =  w\epsilon_\phi(\zt; y) + (1-w)\epsilon_\phi(\zt;\varnothing)) ,
\label{eq:classifierfree}
\end{align}
where $w$ is the guidance weight. The CFG generates class-specific content by increasing the differences between the conditional output $\epsilon_\phi(\zt; y)$ and unconditional output $ \epsilon_\phi(\zt;\varnothing)$. Only using the conditional term $\epsilon_\phi(\zt; y)$ in inference does not generate condition specific content~\cite{ho2021classifier}.

One thing to note is that \textbf{the classifier free guidance is only applied during inference of diffusion models}. Based on Eq.~\ref{eq:energy_score}, the training process of diffusion models solely learns the conditional distribution without CFG (or equivalently with CFG w = 1). Therefore, there is a guidance gap between the diffusion model training and inference.

\subsection{Diffusion Priors in Optimization Problems}

One great property of the diffusion model is that it not only can be used to generate images through denoising, but on a wider range of applications as a diffusion prior. Early work Plug-and-Play Prior~\cite{graikos2022diffusion} treats the diffusion model the same way as other generative methods. To calculate the gradient of the input images, the method back-propagates the gradient through the diffusion U-Net. The milestone work Dreamfusion~\cite{poole2022dreamfusion} experimentally points out that the gradient of U-Net harms the performance of diffusion prior and proposes the SDS loss. As also observed in Score Jacobian Chaining~\cite{wang2023score}, one reason for such property is the score function nature of diffusion models. Therefore, the calculated score can be directly applied to the input image. Specifically, the  SDS loss is defined as the classifier free guidance minus the added noise: 
\begin{align}
\nabla_{\theta} \ldist(\phi, x) = \mathbb{E}_{t, \epsilon}\left[\left(\hat\epsilon_\phi(\zt; y, t)  - \epsilon\right) {\frac{\partial x}{\partial \theta}}\right].
\label{eq:sdsgrad}
\end{align}
To avoid ambiguity in notations, we hide the factor $w(t)$ in SDS loss for simplicity. Although the SDS loss represents a big milestone in using diffusion prior, there are two major limitations: the lack of generation details and diversity. Several methods are proposed to improve the performance of the SDS loss. DDS loss~\cite{hertz2023delta}, proposed for the 2D image editing problem, shows that there exists a noise component in the SDS loss and the noise can be removed by calculating the SDS loss of the source image and text pairs. Concretely, the DDS loss can be calculated by:
\begin{equation}
\begin{aligned}
\nabla_{\theta} \ldistdds(\phi, x&=g(\theta)) =  \mathbb{E}_{t, \epsilon}[((\hat\epsilon_\phi(\zt; y, t) \\
& - \hat\epsilon_\phi(\hat\zt; \hat y, t)) {\frac{\partial x}{\partial \theta}}],
\label{eq:ddsgrad}
\end{aligned}
\end{equation}
where $\zt$ and  $y$ represent the current image and target text description and $\hat\zt$ and  $\hat y$ represent the source image and text description. 

The VSD loss in ProlificDreamer~\cite{wang2023prolificdreamer} is another variant of the SDS loss to improve 3D generation. It learns the variational distribution of the scene through a LoRA model $\hat\epsilon_\psi$ and calculates the diffusion prior by taking the difference between the pretrained model and the LoRA model:
\begin{equation}
\begin{aligned}
\nabla_{\theta} \ldistvsd(\phi,\psi, x&=g(\theta)) =  \mathbb{E}_{t, \epsilon}[(\hat\epsilon_\phi(\zt; y, t) \\
& - (\hat\epsilon_\psi(\zt; y, t)) {\frac{\partial x}{\partial \theta}}].
\label{eq:vsdgrad}
\end{aligned}
\end{equation}

Whereas these methods improve the performance of using diffusion models in optimization problems, the underlying problems of the original SDS loss are still under-explored. Furthermore, there is still a lack of a unified view of how the issues of SDS loss should be resolved. Unless otherwise stated, in this paper, we focus on conditional diffusion models with classifier free guidance, which is the most commonly used diffusion model nowadays.

\section{Method}
\label{sec:method}

\subsection{Problem Formulation}

Consider the problem of optimizing a 3D model~\cite{mildenhall2021nerf} parameterized with parameter $\theta$, and a differentiable rendering operation $g$ transforming $\theta$ into the 2D image $x = g(\theta)$. We are interested in the problem of optimizing $\theta$ with a conditional pretrained diffusion model $\epsilon_\phi(\zt; y)$. In the following subsection, we analyze how this problem is resolved in the SDS loss and why it does not generate good results.

\subsection{Issues of Previous Diffusion Priors}

\textbf{The Divergence in Training and Inference of Diffusion Models Leads to Suboptimal Results with the SDS Loss.} Consider the case of the SDS loss in Eq.~\ref{eq:sdsgrad}. It directly optimizes $\theta$ with a CFG variant  (typically with a weighting factor of 100) of the reference denoising score. However, as mentioned in Eq.~\ref{eq:energy_score}, \textbf{the training of diffusion model learns the score function $\epsilon_\phi(\zt; y, t)$ without employing CFG}. This discrepancy creates a significant issue: the classifier-free guidance applied in the SDS loss does not steer the target distribution to align with the reference distribution, which is characterized by the score function $\epsilon_\phi(\zt; y,t)$, but rather towards a CFG-modified version of the diffusion model, denoted as $\hat\epsilon_\phi$. This leads to the outputs generated by the SDS loss often being overly saturated and lacking in diversity, as noted in the original research~\cite{poole2022dreamfusion}. 

\textbf{A Higher CFG Guidance is Necessary For Diffusion Priors.} A straightforward solution to the previously mentioned problem could be directly removing the CFG in the SDS loss. We term it as \textbf{the reference SDS loss}:
\begin{equation}
\begin{aligned}
\nabla_{\theta}\ldistref(\phi, x) & = \mathbb{E}_{t, \epsilon}[(\epsilon_\phi(\zt; y,t) - \epsilon){\frac{\partial x}{\partial \theta}} ].
\label{eq:sdsgradnorm_final}
\end{aligned}
\end{equation}
However, directly using the above equation is not feasible in 3D generation from both empirical observations and theoretical analysis. Empirically, as witnessed in previous works~\cite{poole2022dreamfusion,lin2023magic3d}, diffusion priors are only capable of learning detailed features of 3D objects when employing large CFG weights $w$. Our experiments have observed similar challenges. An illustration can be found in the experiment section and Fig.\ref{fig:ablation_3d}. Theoretically, \textbf{a large CFG weight $w$ guides the target distribution to learn towards the direction of conditional score function and away from the unconditional score function, thereby generating more content relevant to the condition}. Compared with 2D space, 3D optimization introduces additional out-of-distribution factors~\cite{wang2023score} when using 2D diffusion models. Therefore, the 2D diffusion priors require an even larger $w$ on 3D problems. 

\subsection{Learn to Optimize Denoising Scores}
Based on the preceding analysis, our key insight to improve the diffusion prior is that, the SDS should commence with a high initial classifier-free guidance (CFG) value and eventually align with the reference SDS formula Eq.~\ref{eq:sdsgradnorm_final} to bridge the gap between training and inference stages. A natural approach could be to gradually decrease the CFG weight $w$ during the optimization process. However, this straightforward method did not yield improved results in our experiments. The primary challenge lies in the fact that the CFG weight $w$ is a scalar that uniformly affects the entire noise map, without accounting for internal variations. Moreover, adjusting the $w$ value to suit the varying difficulties of different 3D objects during optimization proves challenging.

To this end, we propose the LODS (Learn to Optimize Denoising Scores) algorithms. Our approach begins by expanding the classifier-free guidance formula with two additional learnable parameters. The first, denoted as $\alpha$ refers to a learnable unconditional embedding initialized with null embedding $\varnothing$. The second $\psi$ refers to additional parameters added to the network (e.g. LoRA~\cite{hu2021lora} parameters). Each of these learnable parameters corresponds to a variant of our proposed method. We then propose to learn the two learnable parameters with the LODS algorithm as illustrated in Algorithm~\ref{algo:algo1}.
\begin{algorithm}[!ht]
\DontPrintSemicolon

Initialize 3D model parameter $\theta$ \;
Initialize the SDS loss with high CFG \;
Initialize $\alpha$ or $\psi$ \;
  \While {Not Converge}
    {   
    Optimize $\theta$ based on SDS with $\alpha$, $\psi$ frozen\;
    Optimize $\alpha$ or $\psi$\ by aligning SDS to Eq.~\ref{eq:sdsgradnorm_final}\;
    } 

\textbf{RETURN} $\theta$
\caption{The LODS Algorithm.}
\label{algo:algo1}
\end{algorithm}

Concretely, we first initialize the 3D model parameters and the current running SDS loss. We then proceed with two iterative optimization steps. In \textbf{Step 5}, we use the current SDS loss to optimize the 3D model parameters. Following this, in \textbf{Step 6}, we optimize the parameters of the SDS. This iterative optimization algorithm offers advantages on two fronts. Firstly, it allows the optimization of the 3D model to commence from arbitrary initial score functions. Secondly, the optimization process in \textbf{Step 6} learns to bridge the gap between the training and inference phases by aligning the original SDS to Eq.~\ref{eq:sdsgradnorm_final}.

The subsequent subsections delve into the specifics of how the LODS algorithm is implemented by optimizing two additional learnable parameters. In this study, we restrict our exploration to the extension of the classifier-free guidance with either a learnable null embedding or learnable low rank parameters. However, it is worth noting that our framework could potentially be expanded to incorporate other learnable parameters, such as those in the ControlNet structure~\cite{zhang2023adding} and the T2I adapter structure~\cite{mou2023t2i}.

\subsection{Embedding Based Optimization}
We define the initial training loss as a normalized version of the SDS loss~\cite{poole2022dreamfusion} with learnable unconditional embedding $\alpha$. Specifically, we use the classifier free guidance weight to normalize the SDS loss:
\begin{equation}
\begin{aligned}
\nabla_{\theta}\ldistnorm (\phi, x&=g(\theta)) = \mathbb{E}_{t, \epsilon}[(\epsilon_{\phi}(\zt; y,t) + \\
& \frac{(1-w)\epsilon_{\phi}(\zt;\alpha,t))}{w}  - \frac{\epsilon}{w}) {\frac{\partial x}{\partial \theta}} ].
\label{eq:sdsgradnorm}
\end{aligned}
\end{equation}
The normalized SDS loss can produce the same results as the original SDS loss by only adjusting the learning rate. Here, we keep the parameters of the conditional distribution $\epsilon_\phi(\zt; y)$ as the original parameters, and the only learnable component is the learnable unconditional embedding $\alpha$. We then learn the learnable unconditional embedding to make the above SDS align to Eq.~\ref{eq:sdsgradnorm_final}. The learnable unconditional embedding can be learned by take the L2 difference (according to Fisher Divergence~\cite{germain2015made}) between Eq.~\ref{eq:sdsgradnorm} and Eq.~\ref{eq:sdsgradnorm_final}. Although the Fisher Divergence describes two score functions, it can be naturally applied to the SDS losses, since the SDS loss is simply the score function minus the random noise. Therefore, the optimization loss for the unconditional embedding $\alpha$ becomes: 
\begin{equation}
\begin{aligned}
 &\parallel \ldistnorm - \ldistref \parallel_2^2 \: \Rightarrow \\
  &\parallel \epsilon_{\phi}(\zt,\alpha,t) - \epsilon \parallel_2^2,
\label{eq:fisherconvert}
\end{aligned}
\end{equation}
if we omit the constant factor. Interestingly, we observe that the calculation of Fisher Distance is independent of the classifier free guidance weight. 

The algorithm is described in Algorithm~\ref{algo:algo2}.

\begin{algorithm}[!ht]
\DontPrintSemicolon
Initialize 3D model parameter $\theta$ \;
Initialize learnable $\alpha$ with null embedding $\varnothing$ \;
Initialize Eq.~\ref{eq:sdsgradnorm} with high CFG \;

  \While {Not Converge}
    {  
    Optimize $\theta$ based on Eq.~\ref{eq:sdsgradnorm} with $\alpha$ frozen\;
    Optimize $\alpha$ based on Eq.~\ref{eq:fisherconvert}\;

    } 

 \textbf{RETURN} $\theta$
\caption{The LODS Algorithm, Update Embedding.}
\label{algo:algo2}
\end{algorithm}

\subsection{Parameter Based Optimization}
It is also possible to optimize the parameters added by a low rank model like LoRA~\cite{hu2021lora}. Similarly, we consider the normalized SDS loss as the initial state:

\begin{equation}
\begin{aligned}
\nabla_{\theta}\ldistnorm(\phi, x&=g(\theta))  = \mathbb{E}_{t, \epsilon}[(\epsilon_\phi(\zt; y,t) + \\
& \frac{(1-w)\epsilon_\psi(\zt;\varnothing,t))}{w}  - \frac{\epsilon}{w}) {\frac{\partial x}{\partial \theta}}],
\label{eq:sdsgradnormlora}
\end{aligned}
\end{equation}
where $\epsilon_\psi$ represents the LoRA model. Here, we still reuse the conditional distribution $\epsilon_\phi(\zt; y)$ of the pretrained diffusion model. Similar to the previous method, the optimization loss for the LoRA parameter $\psi$ is now:
\begin{equation}
\begin{aligned}
 \parallel \epsilon_\psi(\zt;\varnothing,t) - \epsilon \parallel_2^2.
\label{eq:fisherconvertlora}
\end{aligned}
\end{equation}
Finally, the algorithm can be summarized below:
\begin{algorithm}[!ht]
\DontPrintSemicolon
Initialize 3D model parameter $\theta$ \;
Initialize LoRA parameters $\psi$ \;
Initialize Eq.~\ref{eq:sdsgradnormlora} with high CFG \;

  \While {Not Converge}
    {  
    Optimize $\theta$ based on Eq.~\ref{eq:sdsgradnormlora} with $\psi$ frozen\;
    Optimize $\psi$ based on Eq.~\ref{eq:fisherconvertlora}\;

    } 

 \textbf{RETURN} $\theta$
\caption{The LODS Algorithm, Update Parameters.}
\label{algo:algo3}
\end{algorithm}

\begin{figure}[t]
\begin{center}
\includegraphics[trim=0 0 0 0,width=1\linewidth]{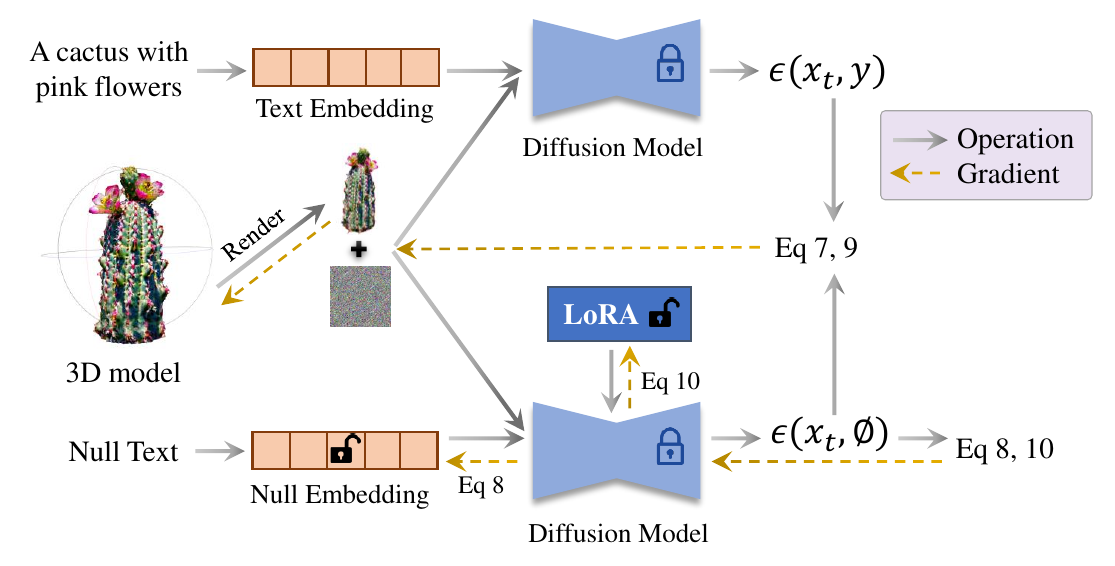}
	\caption{Method overview. Our proposed methods learn to optimize the denoising scores either by optimizing the null embedding or additional low rank parameters of LoRA.}
	\label{fig:model_overview}
\end{center}
\vspace{-0.6cm}  
\end{figure}

Note that both Algorithm~\ref{algo:algo2} and Algorithm~\ref{algo:algo3} are illustrated in the generation case. When used in image editing, the initial state of learnable embedding or the LoRA condition should be set to the source captions. We also show a visual illustration of our proposed methods in Fig.~\ref{fig:model_overview}.

\subsection{Relation with Other Methods}
\textbf{Relation with SDS loss}: As we state previously, the
SDS loss equals we directly optimize the parameter $\theta$ with a CFG version of the reference SDS. In Algorithm~\ref{algo:algo1}, the SDS loss is equivalent to optimizing the \textbf{Step 5} and ignoring \textbf{Step 6}. 

\textbf{Relation with VSD loss}:
Consider the fact that the initial state of LoRA (Hugging Face implementation) is the same as the pretrained model. The initial state of VSD equals our method with infinitely large CFG weight (regardless of the CFG mentioned in VSD). This setup (infinity large weight) leads to the ``floating phenomenon'' in 3D generation. By controlling the initial classifier free guidance, our method can successfully reduce the ``floating phenomenon''. We show more results in the experiment section.

\textbf{Relation with DDS loss}:
Consider the image editing case where the initial state null embedding is set to the source caption. DDS loss is similar to our embedding optimization method without updating the embeddings.

\subsection{Computational Cost Comparison}
\begin{figure}[ht]
\begin{center}
\includegraphics[trim=0 0 0 0,width=1\linewidth]{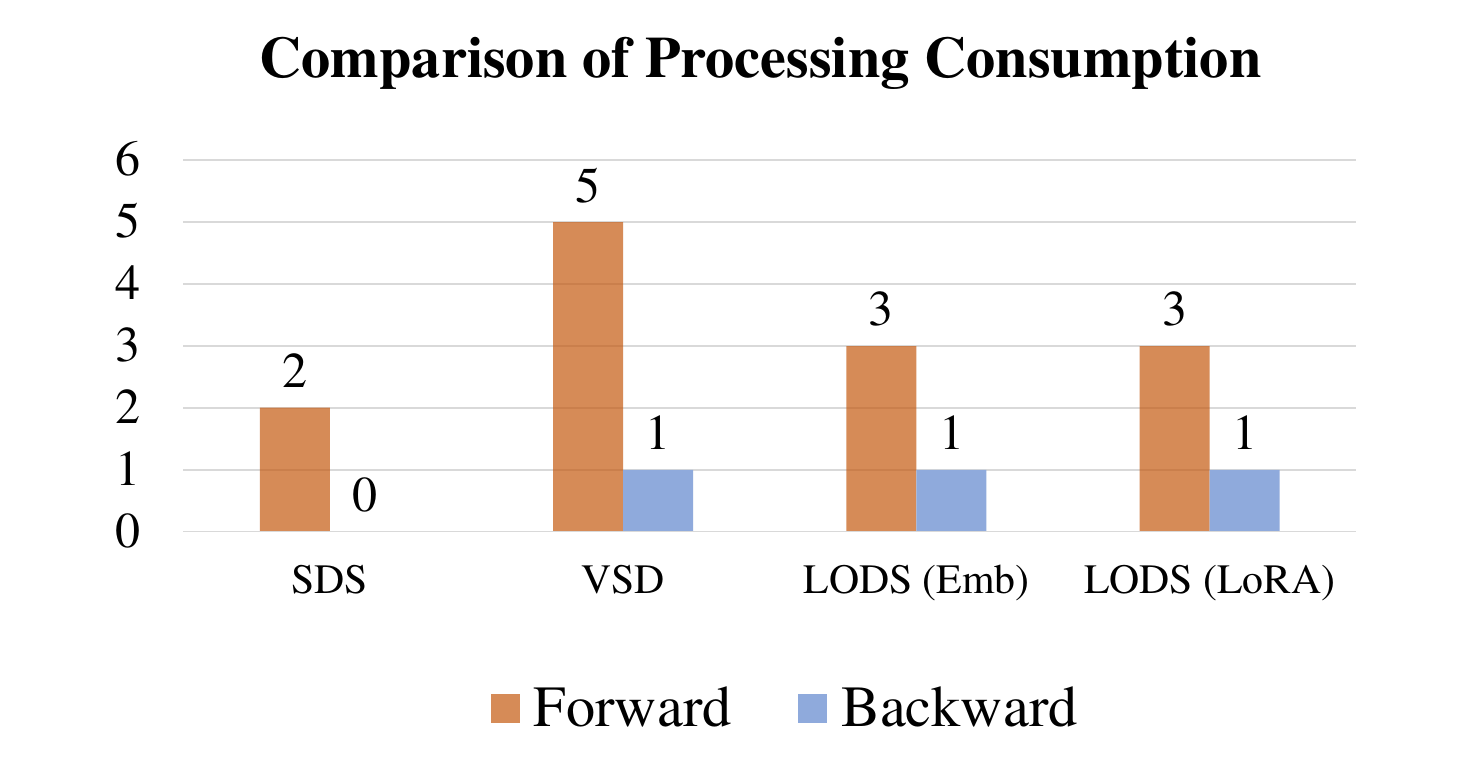}
	\caption{Comparison of processing efficiency. Our proposed methods are more efficient compared with the VSD loss.}
	\label{fig:compare_speed}
\end{center}
\vspace{-0.4cm}  
\end{figure}

The number of required backward and forward diffusion U-Net propagations is illustrated in Fig~\ref{fig:compare_speed}. Compared with the SDS loss baseline, our two proposed methods require one additional forward and backward propagation to calculate the updated denoising score. Compared with the VSD loss, our method demands less computation. 

It is also possible to further improve the efficiency by merging the two conditional forwards of Line 5 and Line 6 in Algorithm~\ref{algo:algo2} and \ref{algo:algo3} into one, such that our method requires only one more backward propagation compared with the SDS loss. However, we notice a bit of the performance drop with this setting. We leave this experiment for future work.

\begin{figure*}[ht]
\begin{center}
\includegraphics[trim=0 0 0 0,width=1\linewidth]{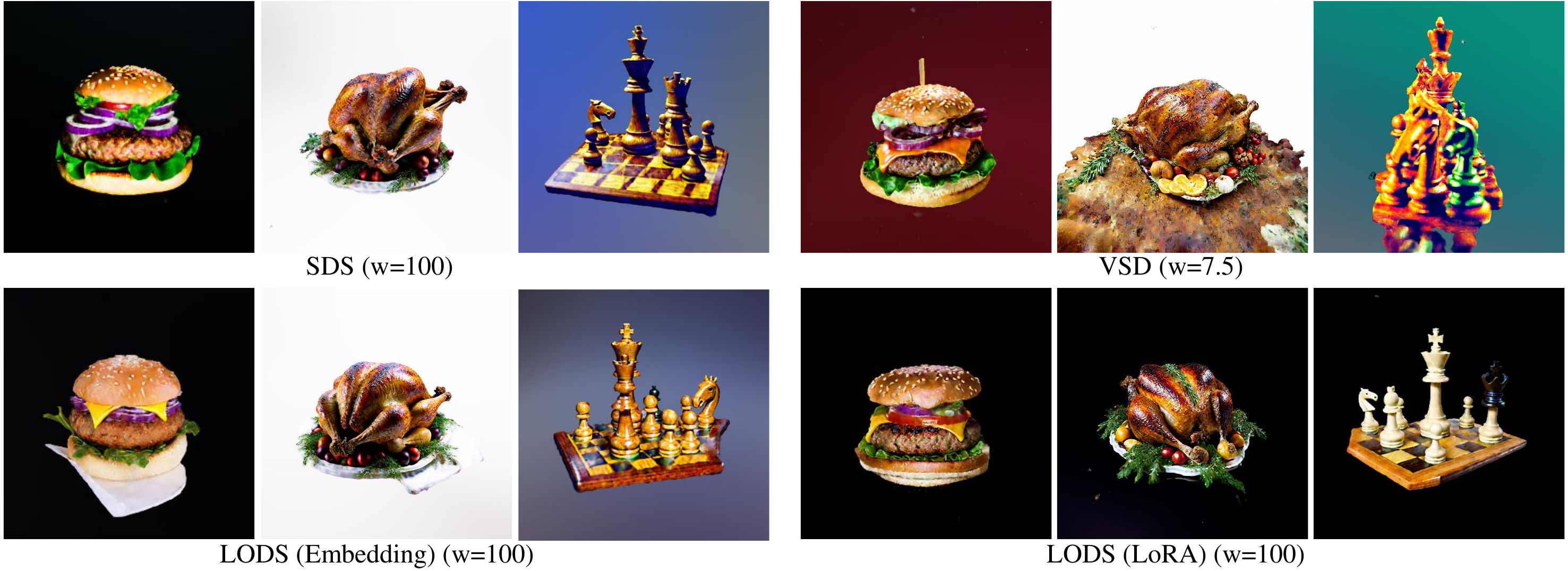}
	\caption{Comparison with other methods on Text-to-3D generation with NeRF. Our method could generate high-quality 3D details compared with the SDS loss and avoid the ``floating'' problem of the VSD loss. Text prompts: ``A DSLR photo of a hamburger'', ``A roast turkey on a platter'', and ``An intricately-carved wooden chess set''. }
	\label{fig:compare_t3d}
\end{center}
\vspace{-0.4cm}  
\end{figure*}
\begin{table*}[ht]
\centering
\caption{Results on Text-to-3D. Our experiments are carried out on T3Bench dataset. Our proposed methods achieve state-of-the-art performance on all testing categories.}
\label{table:text-to-3D}
\resizebox{\textwidth}{!}{%
\begin{tabular}{@{}c|ccccccccc@{}}
\toprule
\textbf{Dataset} & \textbf{Metrics} 
& Dreamfusion~\cite{poole2022dreamfusion} & Magic3D~\cite{lin2023magic3d}
& LatentNeRF~\cite{metzer2023latent} & Fantasia3D~\cite{chen2023fantasia3d}  & SJC~\cite{wang2023score}
& ProlificDreamer~\cite{wang2023prolificdreamer} & \textbf{LODS (Embedding)} & \textbf{LODS (LoRA)} \\\midrule
\multirow{3}{*}{\textbf{Single Object}}        
& Quality$\uparrow$  & 24.9 & 38.7 & 34.2 & 29.2 & 26.3 & 51.1 & \textbf{55.1} & 54.3 \\
&Alignment$\uparrow$             & 24.0 & 35.3& 32.0 & 23.5 & 23.0 & 47.8 & \textbf{49.5} & 48.2 \\
&Average$\uparrow$            & 24.4 & 37.0& 33.1 & 26.4 & 24.7 & 49.4 & \textbf{52.3} & 51.3 \\
\midrule
\multirow{3}{*}{\textbf{Single with Surroundings}} 
& Quality$\uparrow$ & 19.3 & 29.8 & 23.7 & 21.9 & 17.3 & 42.5 & \textbf{42.6} & 40.1\\
& Alignment$\uparrow$         & 29.8 & 41.0& 37.5 & 32.0 & 22.3 & 47.0 & \textbf{57.0} & 54.5 \\
&Average$\uparrow$            & 24.6 & 35.4 & 30.6 & 27.0 & 19.8 & 44.8 & \textbf{49.8} & 47.3 \\
\midrule
\multirow{3}{*}{\textbf{Multiple Objects}} 
& Quality$\uparrow$ & 17.3 & 26.6 & 21.7 & 22.7 & 17.7 & 45.7 & \textbf{46.3} & 43.8 \\
& Alignment$\uparrow$           & 14.8 & 24.8 & 19.5 & 14.3 & 5.8 & 25.8 & \textbf{33.0} & 31.3 \\
& Average$\uparrow$           & 16.1 & 25.7 & 20.6 & 18.5 & 11.7 & 35.8 & \textbf{39.7} & 37.5 \\
\midrule
\multirow{1}{*}{\textbf{Average}} 
& Average$\uparrow$           & 21.7 & 32.7 & 28.1 & 24.0 & 18.7 & 43.3 & \textbf{47.3} & 45.4 \\
\bottomrule
\end{tabular}
}
\end{table*}

\section{Experiments}
In this section, we present comprehensive comparison and ablation experiments in 3D space, encompassing both text-to-3D and image-to-3D generation. Subsequently, we also discuss the applications of our method in 2D space. We include the implementation details of all these experiments in the Appendix for the reference of readers. 
\begin{table*}[ht]
\centering
\caption{Results on Image-to-3D. We show quantitative results in terms of CLIP-Similarity$\uparrow$ / PSNR$\uparrow$ / LPIPS$\downarrow$. Our proposed method improves the baseline method Magic123.}
\label{table:Image-to-3D}
\resizebox{\textwidth}{!}{%
\begin{tabular}{@{}c|cccccccccc@{}}
\toprule
\textbf{Dataset} & \textbf{Metrics} 

& 3DFuse~\cite{seo2023let} & NeuralLift~\cite{xu2023neurallift}  & RealFusion~\cite{melaskyriazi2023realfusion}
& Zero-1-to-3~\cite{liu2023zero} & Magic123~\cite{qian2023magic123} & \textbf{LODS (Embedding)} &\textbf{LODS (LoRA)}\\
\midrule
\multirow{3}{*}{\textbf{NeRF4}}        
& CLIP-Similarity$\uparrow$   & 0.60 & 0.52 & 0.38 & 0.62 & \textbf{0.80} & \textbf{0.80} &0.79 \\
& PSNR$\uparrow$              & 5.86 & 12.55 & 15.37 & 23.96 & 24.62 & \textbf{24.83}&24.67 \\
& LPIPS$\downarrow$            & 0.76 & 0.50 & 0.20 & 0.05 & \textbf{0.03} & \textbf{0.03}&\textbf{0.03} \\ \midrule
\multirow{3}{*}{\textbf{RealFusion15}} 
& CLIP-Similarity$\uparrow$  & 6.28 & 0.65 & 0.67 & 0.75 & \textbf{0.82} & \textbf{0.82} &\textbf{0.82}\\
& PSNR$\uparrow$             & 18.87 & 11.08 & 0.67 & 19.49 & 19.50 & \textbf{19.63}&19.53  \\
& LPIPS$\downarrow$           & 0.80 & 0.53 & 0.14 & 0.11 & 0.10& \textbf{0.09}& 0.10  \\ 
\bottomrule
\end{tabular}
}
\end{table*}

\begin{figure*}[ht]
\begin{center}
\includegraphics[trim=0 0 0 0,width=1\linewidth]{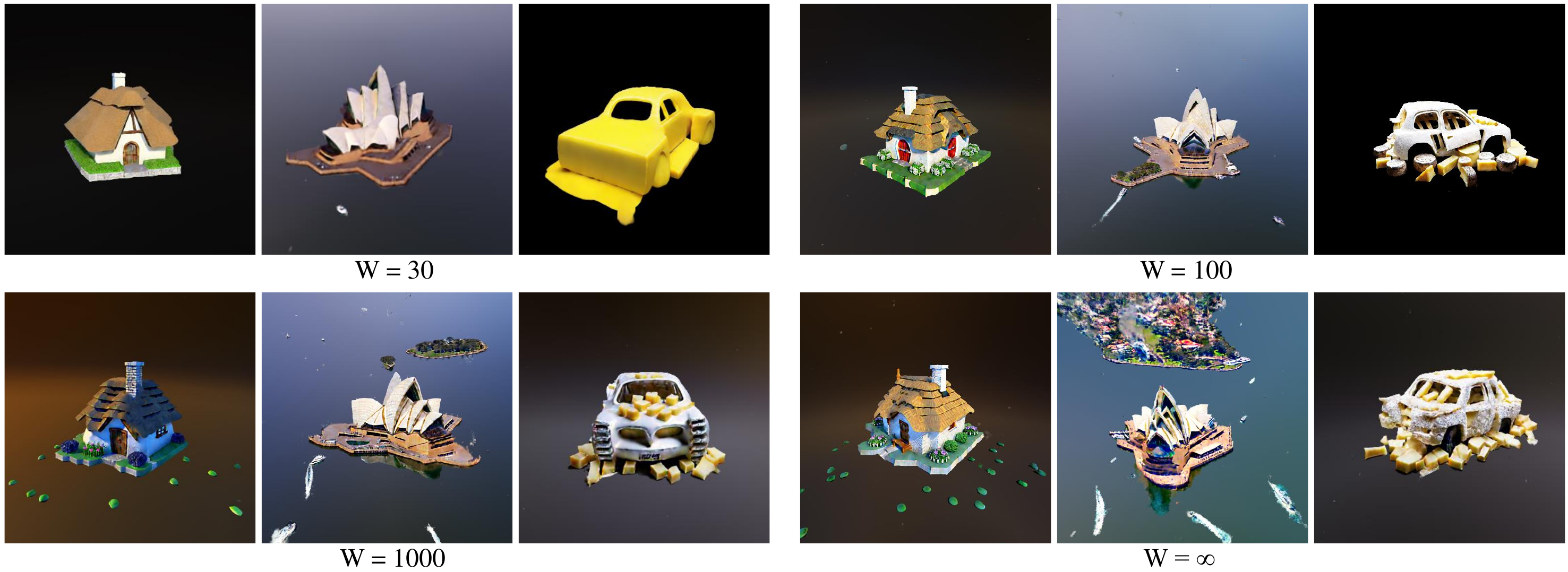}
	\caption{Ablation experiments of the classifier free guidance weight on text-to-3D generation with NeRF. At low CFG weight, the generated results usually lack texture details. When the CFG weight is excessively increased, it leads to a similar ``floating'' phenomenon as observed in VSD. Text prompts: ``A 3D model of an adorable cottage with a thatched roof'', ``Sydney opera house'', and ``A car made out of cheese''.}
	\label{fig:ablation_3d}
\end{center}
\vspace{-0.6cm}  
\end{figure*}

\subsection{Text-to-3D Generation using NeRF backbone}

\subsubsection{Comparison with Other Diffusion Priors.} 
We follow the same text-to-3D generation process as previous methods. For each text prompt, we start from a randomly initialized NeRF and optimize the NeRF parameters with the diffusion priors. We present both qualitative and quantitative results of text-to-3D generation. 

\begin{figure*}[ht]
\begin{center}
\includegraphics[trim=0 0 0 0,width=1\linewidth]{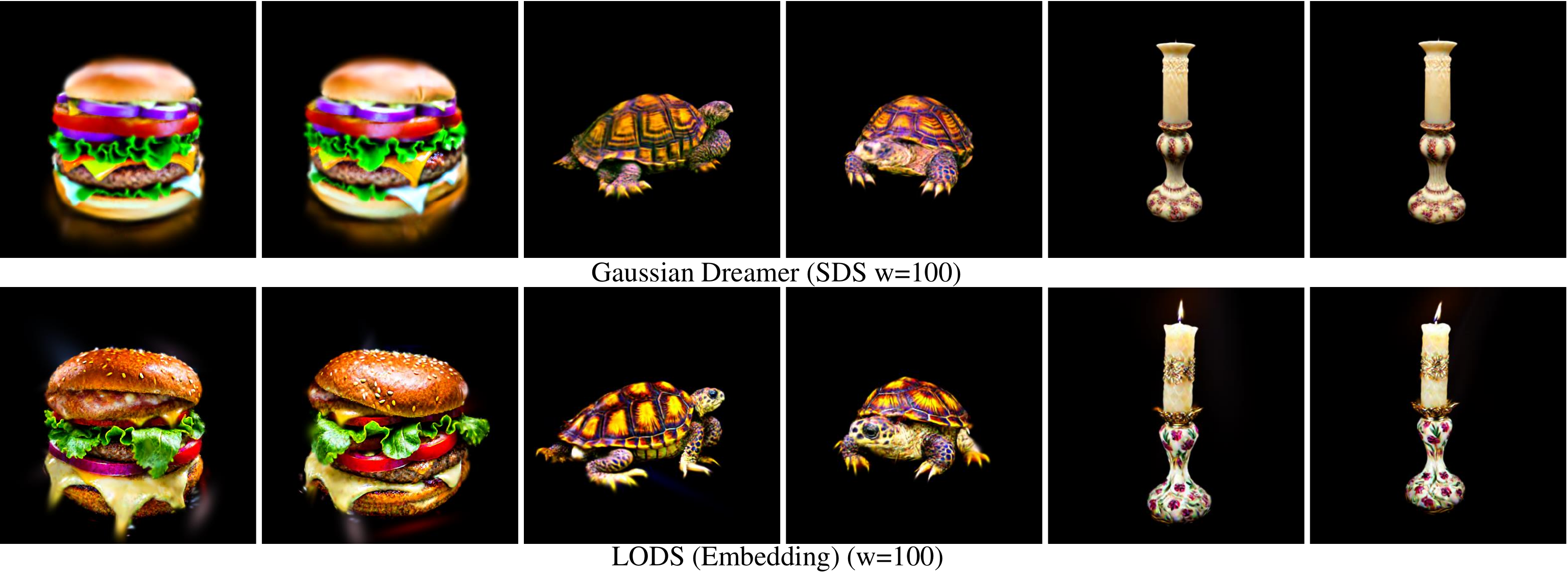}
	\caption{Comparison on Gaussian Splatting. Our method could generate 3D details of superior quality compared with the SDS loss. Text prompts: ``A DSLR photo of a hamburger'', ``A partly broken shell of a tortoise'', and ``An ivory candlestick holder''. }
	\label{fig:compare_gs}
\end{center}
\vspace{-0.4cm}  
\end{figure*}
\begin{figure}[ht]
\begin{center}
\includegraphics[trim=0 0 0 0,width=1\linewidth]{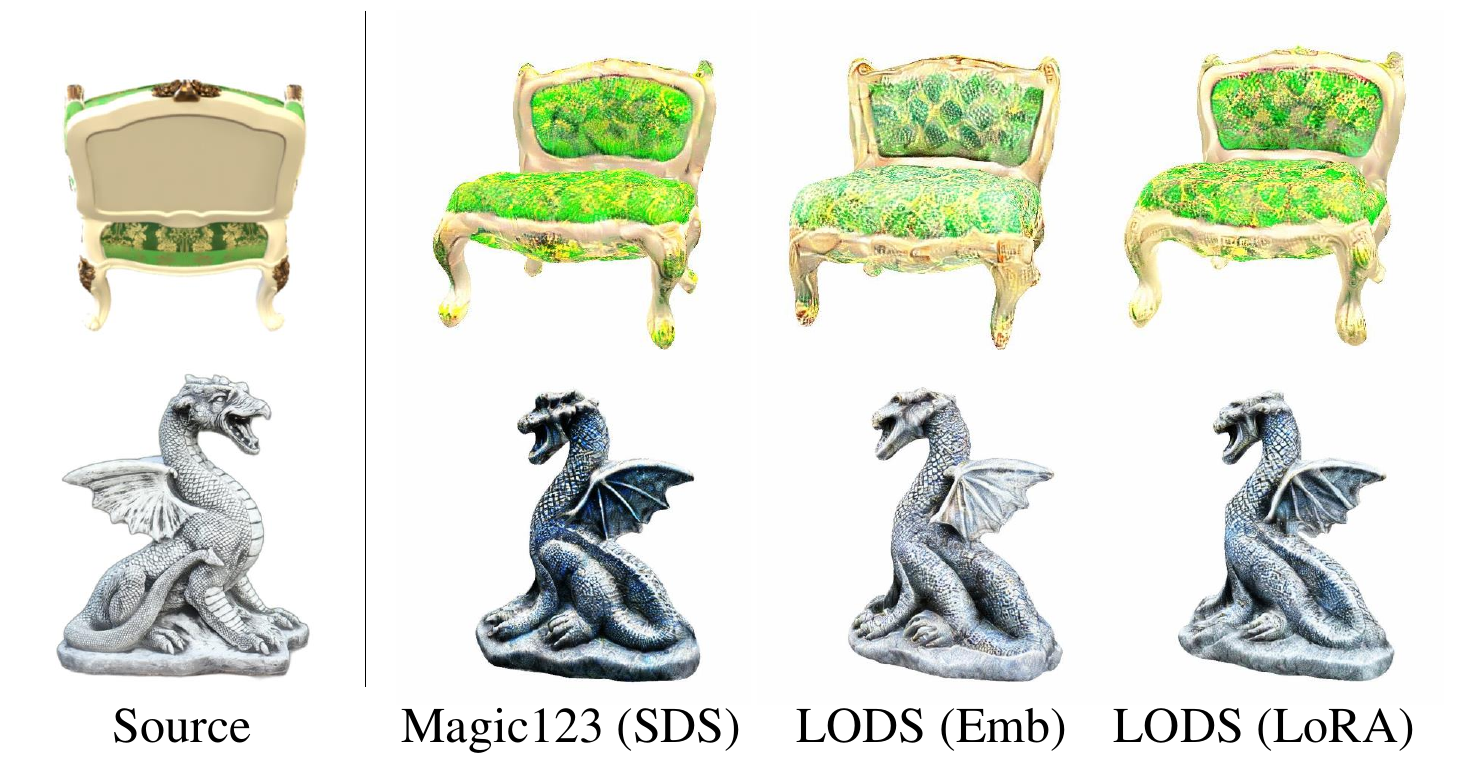}
	\caption{Image-to-3D Generation with NeRF. Our methods are able to generate more texture-consistent novel views.}
	\label{fig:compare_i3d_generation}
\end{center}
\vspace{-0.6cm}  
\end{figure}
\textbf{Qualitative Results.} In Fig.~\ref{fig:compare_t3d}, we display three representative examples generated by our methods in comparison to the SDS loss and the VSD loss. Compared with the SDS loss, both of our methods, as well as the VSD loss, are capable of generating 3D objects with more intricate details (such as the vegetables in a hamburger) and less saturated colors. While the VSD loss can also produce highly detailed objects, it often results in the creation of ``floating'' objects, which considerably diminishes the overall quality of the generation. Our methods successfully circumvent this ``floating'' issue, while simultaneously achieving high-quality detail generation. Additional comparative results are available in the Appendix. We also provide the generated videos in the supplementary materials.

\textbf{Quantitative Results.} To substantiate the efficacy of our proposed methods, we test our methods on the recently released text-to-3D benchmark T3Bench~\cite{he2023t}. The benchmark includes three categories of text prompts, the single object, the single object with surroundings, and multiple objects, each containing 100 text prompts. This sample size is sufficiently large to ensure an accurate assessment. The evaluation metrics include the multi-view ImageReward~\cite{xu2023imagereward} scores and the 3D captioning GPT4 correspondence scores. For consistency and fairness in comparisons, all methods employ the standard text-to-image diffusion models~\cite{rombach2022high} for performing the diffusion prior. According to the T3Bench results as shown in Table.~\ref{table:text-to-3D}, our methods significantly outperform the VSD loss in ProlificDreamer~\cite{wang2023prolificdreamer} and achieve new state-of-the-art performance in text-to-3D generation. 

In addition to quantitative analysis, we conduct a user study to further validate the effectiveness of our methods from the perspective of users. The detailed results of this study are available in the Appendix.

\begin{figure}[ht]
\begin{center}
\includegraphics[trim=0 0 0 0,width=1\linewidth]{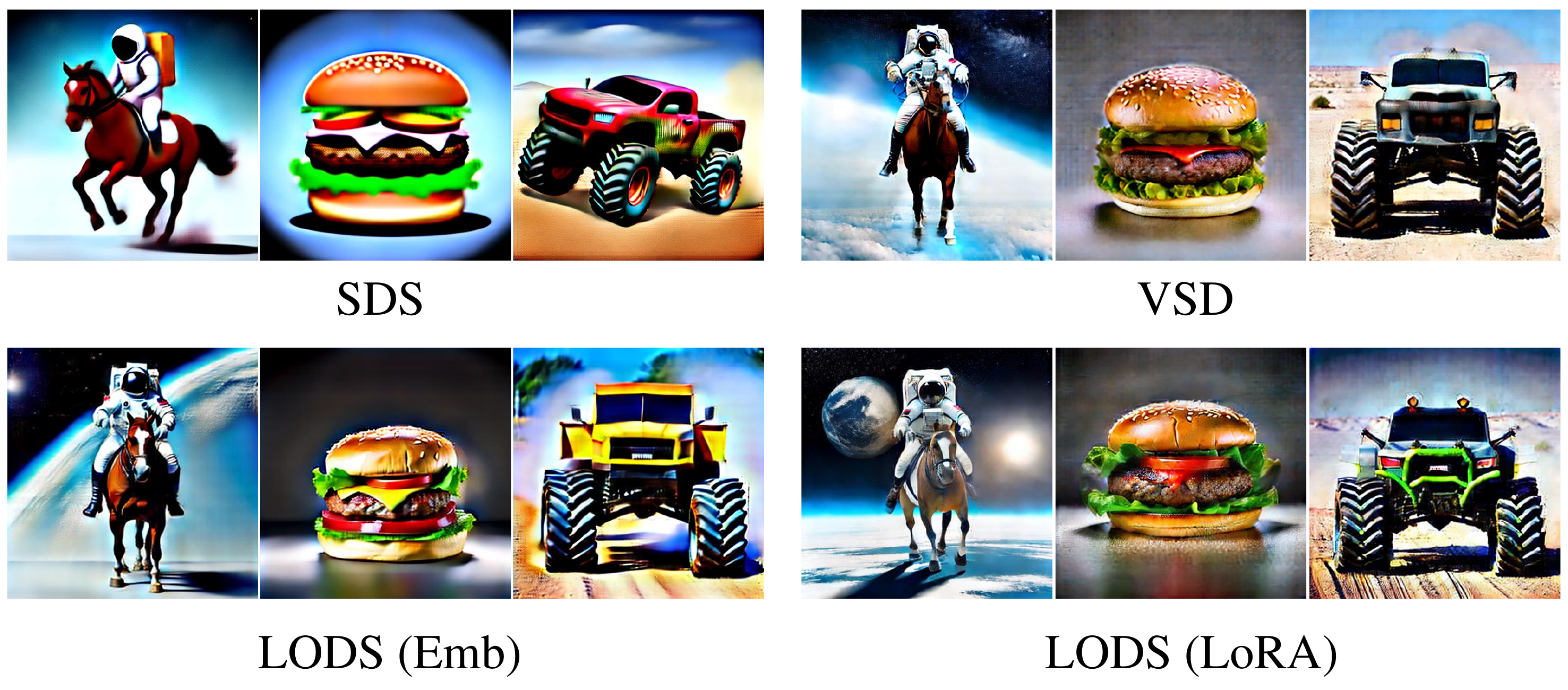}
	\caption{Generation in 2D space. Text-to-image generation using diffusion prior. Our method could generate photo-realistic 2D generation results. The text prompts from left to right are: ``An astronaut is riding a horse'', ``A hamburger'', and ``A monster truck''.}
	\label{fig:compare_2d_generation}
\end{center}
\vspace{-0.4cm}  
\end{figure}
\begin{figure}[ht]
\begin{center}\resizebox{1\linewidth}{!}{
\includegraphics[trim=0 0 0 0,width=1\linewidth]
{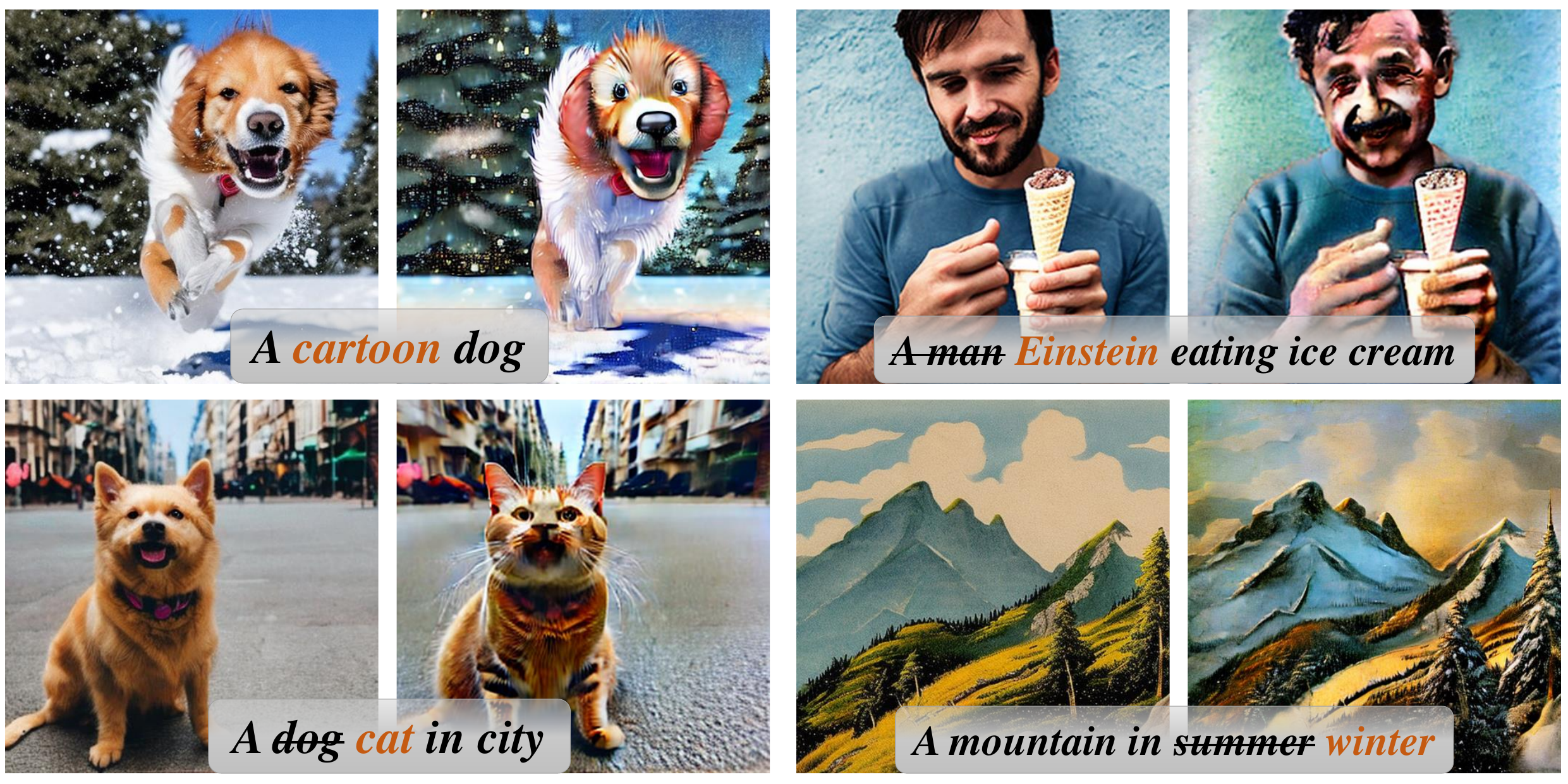}}
	\caption{2D editing. Our diffusion prior can achieve text guided zero-shot editing on 2D images.}
	\label{fig:2d_editing}
\end{center}
\vspace{-0.6cm}  
\end{figure}

\subsubsection{Ablation Experiments on CFG Weights.} 
We show the ablation experiments in Fig.~\ref{fig:ablation_3d}. In this experiment, we gradually increase the initial classifier free guidance (CFG) weight from 30 to infinity large using the embedding-based method. We observe that with low starting CFG weight, the learned results lack details. While for large CFG weights (usually beyond 1000), the generated 3D content tends to generate the ``floating'' objects. This observation explains why the VSD loss usually suffers from the same ``floating'' issue, because the initial state of VSD equals our infinitely large CFG case. We observe very similar behaviors when using our LoRA-based method. We include more ablation experiments in the Appendix.

\subsection{Image-to-3D Generation}
We also show our experimental results of image-to-3D generation. Our method roughly follows the implementation of Magic123~\cite{qian2023magic123}. In Magic123, the authors use two diffusion priors to solve the image-to-3D problem --- one stable diffusion~\cite{rombach2022high} prior to improve texture and one Zero-123 prior~\cite{liu2023zero} to improve shapes. Both of the diffusion priors are conducted with the SDS loss. In experiments, we replace the SDS loss for stable diffusion with our proposed methods and keep the Zero-123 prior as the original SDS loss due to GPU memory limitations. As we illustrate qualitatively in Fig.~\ref{fig:compare_i3d_generation}, the baseline SDS loss often generates over-saturated colors, while our methods can generate more natural novel views compared with the baseline method. We also compare our methods against other image-to-3D methods on the NeRF4 dataset and the RealFusion15 dataset in Table~\ref{table:Image-to-3D}. We observe that our methods achieve the best performance in terms of the CLIP similarity, PSNR, and LPIPS scores.

\subsection{Text-to-3D Generation with Gaussian Splatting}
Our methods extend beyond the NeRF backbone. 3D Gaussian Splatting~\cite{kerbl20233d} is an emerging method for 3D scene representation that utilizes Gaussian points for an explicit and detailed portrayal. This technique facilitates exceptionally rapid learning and rendering. When implemented with 3D Gaussian Splatting, our approach significantly enhances generation speed, achieving a fourfold increase compared to the NeRF backbone (reducing time from 2 hours to just 30 minutes on NVIDIA A6000 GPU). In our experiments, we benchmark our method against the contemporary GaussianDreamer~\cite{yi2023gaussiandreamer}, which employs SDS loss. The comparative results, as illustrated in Fig~\ref{fig:compare_gs}, clearly show that our method excels in producing 3D objects with more lifelike colors and intricate texture details.

\subsection{Generation and Editing in 2D Space}
Although our primary focus is on 3D problems, we also present qualitatively results in 2D space to further demonstrate the effectiveness of our proposed method. In an ideal scenario, these 2D generation experiments are considered as the upper bound for 3D generation. This is achieved by eliminating the rendering function $g$ and directly setting $x = \theta$. The generation results are shown in Fig.~\ref{fig:compare_2d_generation}. Our observations indicate that our methods are capable of generating photo-realistic 2D images. Compared with the SDS loss, our results showcase enhanced object details and more accurate color representation. Our generation results are comparable with the VSD loss in 2D space.

Additionally, our method can be applied to zero-shot 2D editing, as depicted in Fig.~\ref{fig:2d_editing}. An ablation study included in the Appendix further supports the superiority of our method over the SDS loss and other diffusion priors. While our main target remains 3D tasks, future work could explore enhancing 2D editing performance by integrating our proposed diffusion priors with feature injection~\cite{hertz2022prompt,mokady2023null}, textual inversion~\cite{gal2022image,mokady2023null} and model finetuning~\cite{kawar2023imagic}.

\subsection{Discussion of the Two Proposed Methods.} Our two proposed methods exhibit comparable processing speeds. The primary advantage of the embedding-based method lies in its ease of application to newly developed models. This method requires minimal understanding of the internal model structure and can be implemented with less than 20 additional lines of code. This feature is especially beneficial for recent geometry-aware diffusion methods, such as MVDreamer~\cite{shi2023MVDream} and other methods~\cite{liu2023zero,shi2023zero123++} that change the original U-Net structure. The LoRA-based method has shown to produce exceptional results in 2D space. However, in the context of 3D space, the LoRA model sometimes tends to overfit, which can adversely affect the quality of generation. Additionally, our experiments did not reveal any significant benefits from combining the embedding-based and LoRA-based methods. In fact, integrating these two approaches often results in longer training durations, increased memory requirements, and generally fails to yield results that are as effective as those achieved by other methods.
\section{Conclusions, Discussion, and Limitations}
\label{sec:conclusion}
In this paper, we introduce a unified and enhanced diffusion prior for 3D generation, achieving new state-of-the-art performance in text-to-3D generation. Our method also demonstrates the capability to smoothly transition to 2D problems.

\textbf{Limitations.} Currently, our methods are primarily focused on improving the texture aspect of 3D generation. However, they fall short in enhancing the geometry of the generated 3D models.

\textbf{Future Work.} As a versatile diffusion prior, our method holds the potential for seamless integration with other geometry-aware diffusion models~\cite{liu2023zero,shi2023zero123++,ye2023consistent,li2023sweetdreamer,long2023wonder3d,shi2023MVDream}. Future work can be done by combining our method with such methods to further improve the geometric quality. 

{\small
\bibliographystyle{ieeenat_fullname}
\bibliography{11_references}
}

\ifarxiv \clearpage \appendix \section{Appendix Section}
\subsection{Implementation Details of the Experiments}
Our implementations are mainly based on the codebase from ThreeStudio~\cite{threestudio2023}. For all experiments, we use publicly available Stable Diffusion model version 2.1. We use NVIDIA A6000 GPUs with 48G memory.

\textbf{3D Experiments with NeRF.} We use Instant-NGP as the NeRF backbone and the similar implementation hyper-parameters as the Prolific Dreamer configurations. 

\textbf{Experiments on T3Bench:} The embedding based method uses CFG weight 1000 and embedding learning rate 10-5 to update the learnable embeddings. The LoRA based method uses CFG weight 1000, LoRA weight 0.5 and LoRA learning rate 5x10-7. When running on A6000 GPUs, Prolific Dreamer takes around 3 hours to generate one scene, while both of our methods take only less than 2 hours.

\textbf{Image-to-3D Experiments:} We use exactly the same hyper-parameters as Magic123 with CFG 100. The embedding learning rate is set to 10-3. The LoRA learning rate is set to 5x10-7.

\textbf{3D Experiments with 3D Gaussian Splatting.} We briefly follow the 3D Gaussian Splatting implementation of GaussianDreamer. We use Shape-E to initialize the 3DGS. The images are rendered at a size of 512x512.

\textbf{2D Experiments.} We optimize the latent map and use the stable diffusion decoder to decode the learned latent map after the optimization finishes. We use CFG guidance weight infinity. The latent map are optimized with our diffusion priors for 6000 steps with learning rate 3x10-2. For embedding based method, the learnable embedding is updated with a learning rate of 1x10-4. For our LoRA based method, the LoRA parameters are updated with a learning rate of 1x10-6. 

\subsection{Ablation Experiments on 2D Editing}
We show 2D editing ablation results in Fig.~\ref{fig:ablation_2d}. We observe that the original SDS loss generates suboptimal results with blurred images. Our methods could perform effective editing and keep the irrelevant content unchanged at the same time. Based on the visulization, our methods are better than VSD loss and similar to the DDS loss when used on image editing problems.
\begin{figure*}[ht]
\begin{center}
\includegraphics[trim=0 0 0 0,width=1\linewidth]{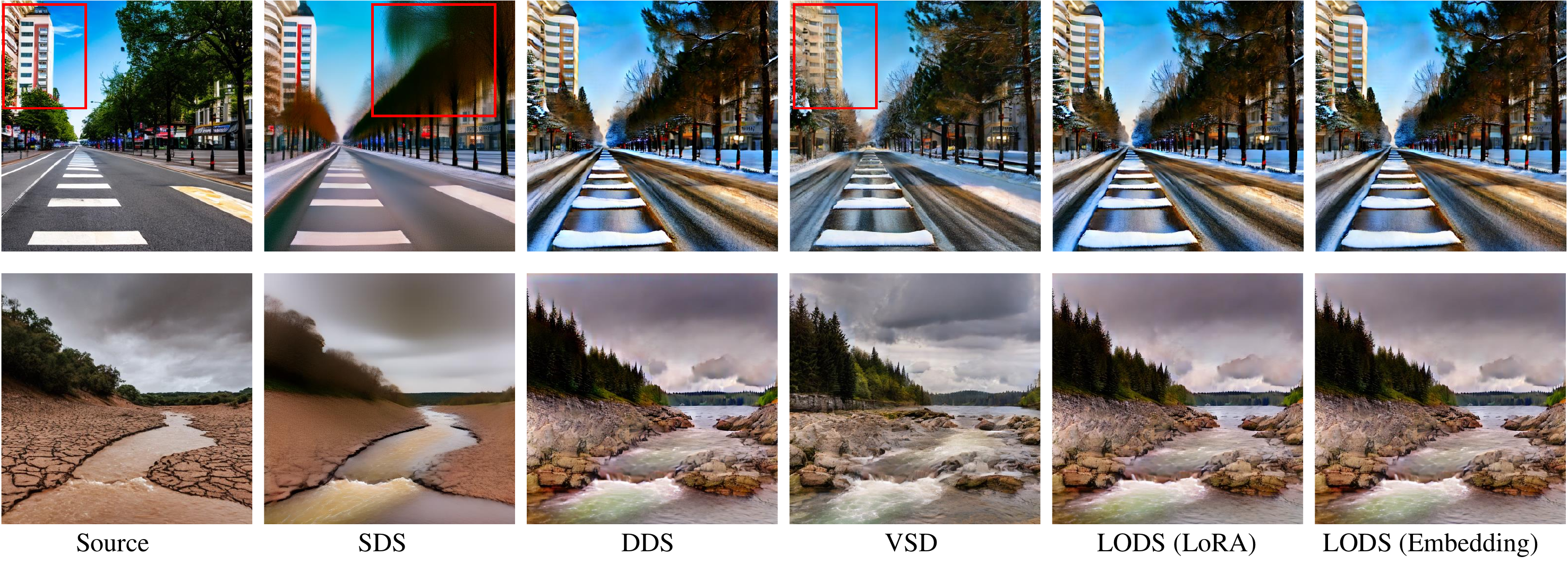}
	\caption{Ablation experiments of 2D editing. Target prompts: Top: ``A DSLR image of a boulevard'' to ``A DSLR image of a boulevard in winter''. Bottom: ``A DSLR image of a dried river'' to ``A DSLR image of a river''. Our proposed method achieves better results compared with the SDS loss and VSD loss.}
	\label{fig:ablation_2d}
\end{center}
\end{figure*}

\subsection{Generation Diversity}
We use generation at 2D space as an example to show that our method can generate diversified content. Results can be found in Fig.~\ref{fig:diversity}.
\begin{figure*}[ht]
\begin{center}
\includegraphics[trim=0 0 0 0,width=1\linewidth]{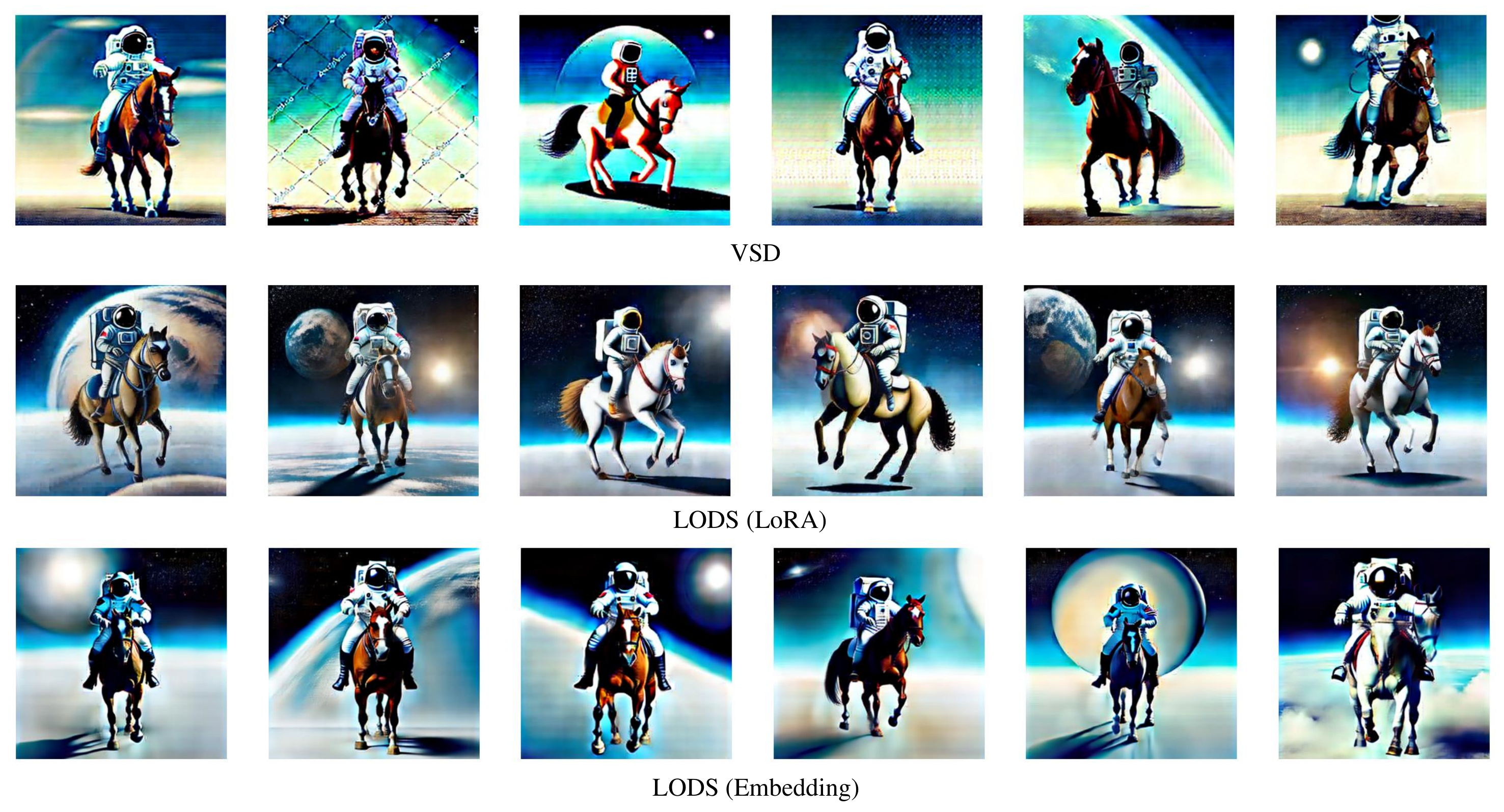}
	\caption{Our method can generate diversified contents. Text Prompt: ``An astronaut is riding a horse''.}
	\label{fig:diversity}
\end{center}
\end{figure*}

\subsection{More 3D Generation Results}
We show more generation results in Fig.~\ref{fig:t3d_more} and Fig.~\ref{fig:t3d_more2}.
\begin{figure*}[ht]
\begin{center}
\includegraphics[trim=0 0 0 0,width=1\linewidth]{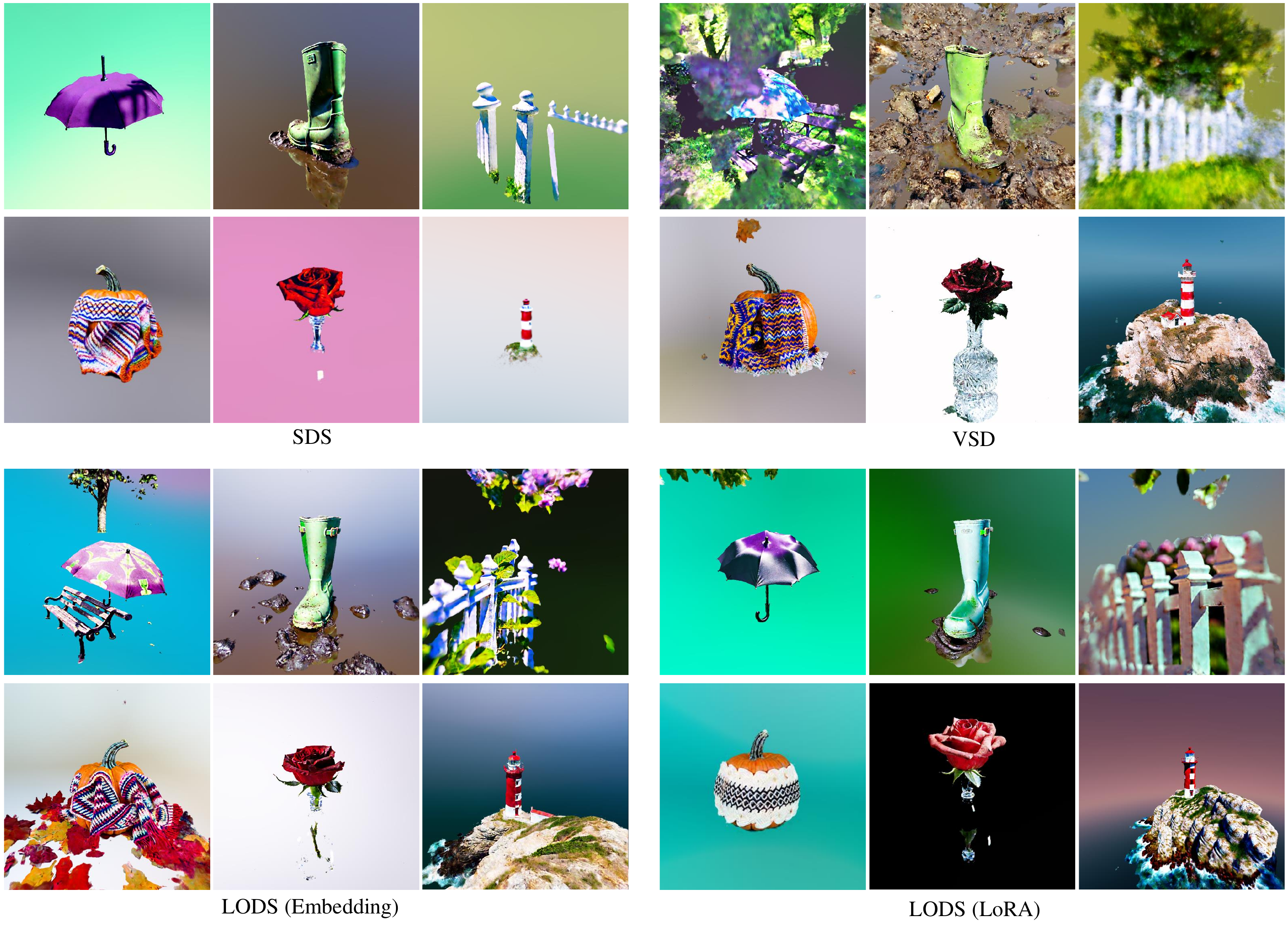}
	\caption{More 3D generation comparison between our proposed methods and other methods with NeRF backbone. Text Prompts: ``A purple umbrella left on a park bench'', ``A green wellington boot in mud'', ``A white picket fence around a garden'', ``A woolen knitted scarf is wrapped around a carved pumpkin'', ``A red rose in a crystal vase'' and ``A red and white lighthouse on a cliff''.}
	\label{fig:t3d_more}
\end{center}
\end{figure*}
\begin{figure*}[ht]
\begin{center}
\includegraphics[trim=0 0 0 0,width=1\linewidth]{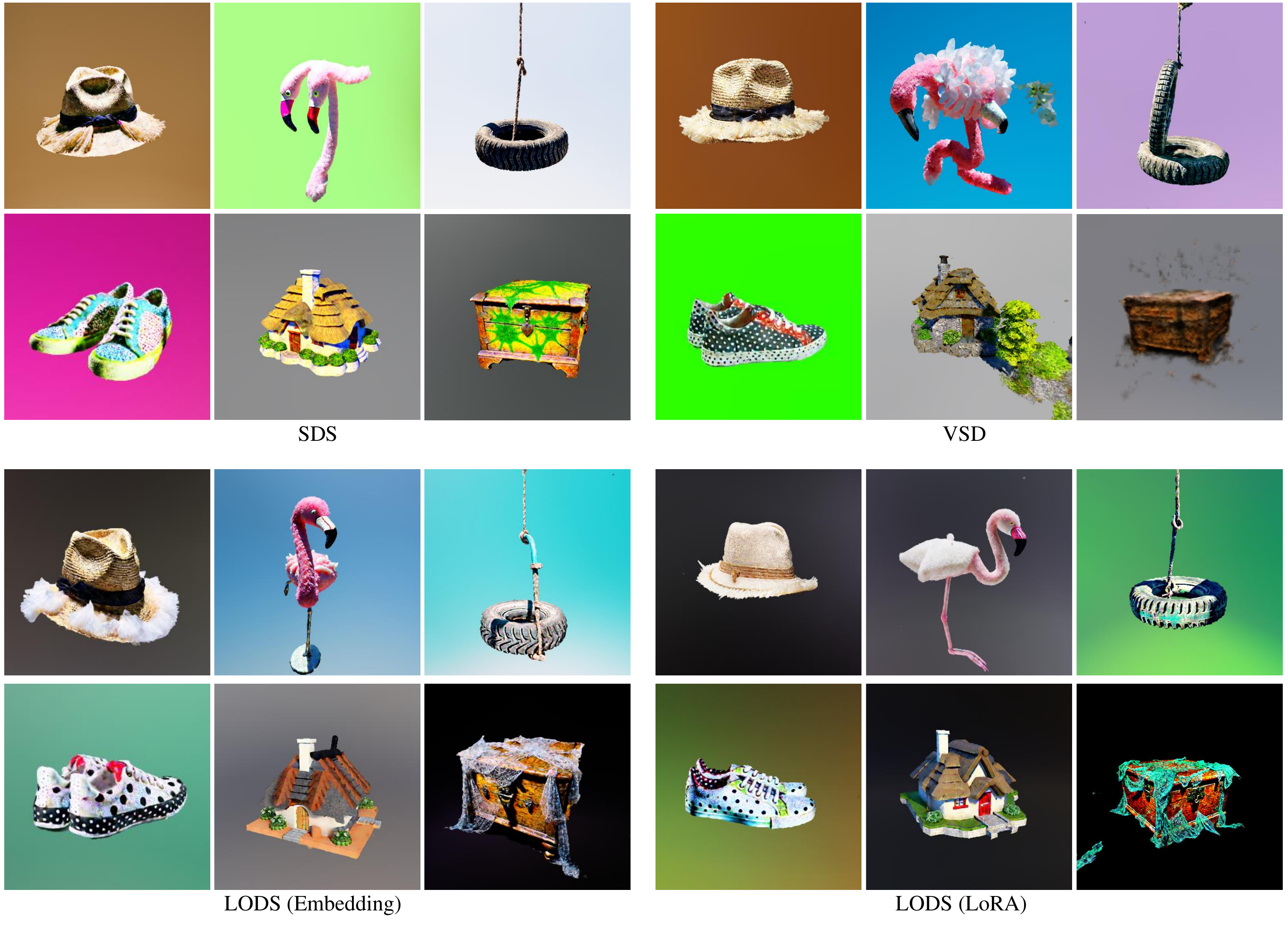}

	\caption{More 3D generation comparison between our proposed methods and other methods with NeRF backbone. Text Prompts: ``An old, frayed straw hat'', ``A fuzzy pink flamingo lawn ornament'', ``A worn-out rubber tire swing'', ``A pair of polka-dotted sneakers'', ``A 3d model of an adorable cottage with a thatched roof'' and ``A cobweb-covered old wooden chest''.}

	\label{fig:t3d_more2}
\end{center}
\end{figure*}

\subsection{More 3D Ablation Experiments}
We show more ablation results in Fig~\ref{fig:ablation_more}.  
\begin{figure*}[ht]
\begin{center}
\includegraphics[trim=0 0 0 0,width=1\linewidth]{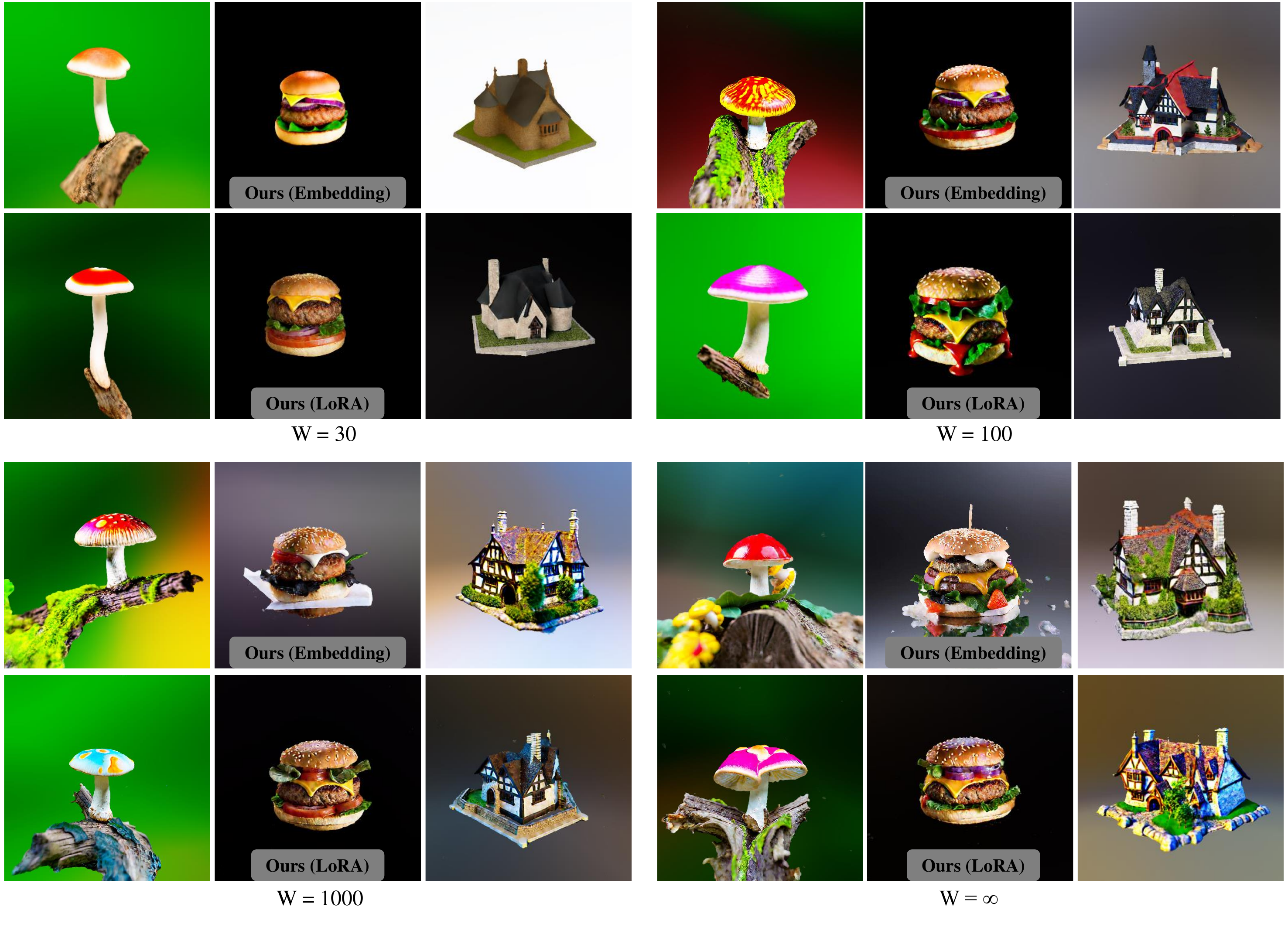}
	\caption{More ablation results of 3D generation using both of our methods with NeRF backbone. Text Prompts: ``A brightly colored mushroom growing on a log'', ``A DSLR photo of a hamburger'' and ``A model of a house in Tudor style''.}
	\label{fig:ablation_more}
\end{center}
\end{figure*}

\subsection {User Study.} 
To further prove the effectiveness of our proposed methods, we conduct user studies to evaluate the users' preferences on the generation results. We use 40 prompts randomly sampled from the T3bench dataset. For each prompt, we generate 3D models with the SDS loss, VSD loss, our embedding, and our LoRA methods. The users are asked to select the best one. Based on the user feedback, the participants are in favour of our embedding based methods on 45\% of the prompts and our lora based methods on 22.5\% of the prompts. The VSD loss and SDS win on 22.5\% of the prompts and 10\% of the prompts respectively. The user study results are consistent with the results we obtained from T3Bench.

\subsection{Why we use the normalized SDS loss as the initial loss}
The use of the normalized SDS loss (Eq.~\ref{eq:sdsgradnorm}) is based on two considerations. Firstly, as mentioned in the main text, the normalized SDS loss produces the same results as the original SDS loss. Secondly, the normalization process makes sure that this initial loss is closer to the reference SDS loss (Eq.~\ref{eq:sdsgradnorm_final}), because the conditional terms are at the same scale. Therefore, the normalized SDS loss greatly eases the learning process toward Eq.~\ref{eq:sdsgradnorm_final}. \fi

\end{document}